\documentclass[lettersize,journal]{IEEEtran}

\usepackage{amsmath,amsfonts}
\usepackage{algorithmic}
\usepackage{algorithm}
\usepackage{array}
\usepackage[caption=false,font=normalsize,labelfont=sf,textfont=sf]{subfig}
\usepackage{textcomp}
\usepackage{stfloats}
\usepackage{url}
\usepackage{verbatim}
\usepackage{graphicx}
\usepackage{cite}
\usepackage[table,xcdraw]{xcolor}
\usepackage{cleveref}
\usepackage{soul}
\usepackage{hhline}

\crefformat{equation}{(#2#1#3)}
\crefmultiformat{equation}{(#2#1#3)}{ and~(#2#1#3)}{, (#2#1#3)}{ and~(#2#1#3)}

\usepackage{mwe,tikz}
\usepackage{cite}
\usepackage{amsmath,amssymb,amsfonts}
\usepackage{algorithmic}
\usepackage{graphicx}
\usepackage{textcomp}
\usepackage{hhline}
\usepackage{tabularx}
\usepackage{soul}
\usepackage[normalem]{ulem}
\usepackage[table,xcdraw]{xcolor}
\usepackage{threeparttable}
\usepackage{booktabs}
\usepackage{array}
\usepackage{mwe,tikz}

\begin{document}


\title{NDST: Neural Driving Style Transfer for Human-Like Vision-Based Autonomous Driving}

\author{
        Donghyun Kim,
        Aws Khalil, 
        Haewoon Nam,~\IEEEmembership{Senior Member,~IEEE}, and
        Jaerock Kwon, \IEEEmembership{Senior Member, IEEE}
        \thanks{A. Khalil and J. Kwon are with the University of Michigan-Dearborn, Michigan, 48128 USA.}
        \thanks{D. Kim and H. Nam are with the Department of Electrical and Electronic Engineering, Hanyang University, Ansan, 15588, South Korea (e-mail: hnam@hanyang.ac.kr).}
        }

\markboth{Journal of \LaTeX\ Class Files,~Vol.~14, No.~8, August~2021}%
{Shell \MakeLowercase{\textit{et al.}}: A Sample Article Using IEEEtran.cls for IEEE Journals}

\IEEEpubid{0000--0000/00\$00.00~\copyright~2021 IEEE}

\maketitle

    \begin{abstract}

Autonomous Vehicles (AV) and Advanced Driver Assistant Systems (ADAS) prioritize safety over comfort. The intertwining factors of safety and comfort emerge as pivotal elements in ensuring the effectiveness of Autonomous Driving (AD). Users often experience discomfort when AV or ADAS drive the vehicle on their behalf. Providing a personalized human-like AD experience, tailored to match users’ unique driving styles while adhering to safety prerequisites, presents a significant opportunity to boost the
acceptance of AVs. This paper proposes a novel approach, Neural Driving Style Transfer (NDST), inspired by Neural Style Transfer (NST), to address this issue.
NDST integrates a Personalized Block (PB) into the conventional Baseline Driving Model (BDM), allowing for the transfer of a user's unique driving style while adhering to safety parameters. The PB serves as a \textit{self-configuring} system, learning and adapting to an individual's driving behavior without requiring modifications to the BDM. This approach enables the personalization of AV models, aligning the driving style more closely with user preferences while ensuring baseline safety critical actuation.
Two contrasting driving styles (Style A and Style B) were used to validate the proposed NDST methodology, demonstrating its efficacy in transferring personal driving styles to the AV system. Our work highlights the potential of NDST to enhance user comfort in AVs by providing a personalized and familiar driving experience. The findings affirm the feasibility of integrating NDST into existing AV frameworks to bridge the gap between safety and individualized driving styles, promoting wider acceptance and improved user experiences.

\end{abstract}

\begin{IEEEkeywords}
AV Personalization, Deep learning, Driving Style, Style Transfer
\end{IEEEkeywords}

    \section{Introduction}
\label{sec:introduction}

In the last decade, Autonomous Vehicles (AV) and Advanced Driver Assistant Systems (ADAS) have progressed significantly, moving closer to their goal of providing a safer and more regulated mode of transportation. However, the attainment of safety alone does not ensure the success of Autonomous Driving (AD) technology \cite{ling2021towards}. Persistent reservations and concerns among the public \cite{dillen2020keep} highlight the critical need for AD technology to not only prioritize safety but also deliver a comfortable and stress-free user experience.

The intertwining factors of safety and comfort emerge as pivotal elements in ensuring the effectiveness of AD. Passenger discomfort arises when a driving style significantly differs from what individuals are accustomed to \cite{ittner2020discomfort}. 
This driving style, including turning dynamics, speed modulation, acceleration, and braking, is a key factor influencing the overall user acceptance of AD technology. Therefore, adopting a human-like driving style is crucial for improving both comfort and safety in the AD experience \cite{vasile2023influences}.

\begin{figure}[!t]
    \centering
    \subfloat[\label{fig:neural_image_style_transfer}]{	  
        \includegraphics[width=1\columnwidth]{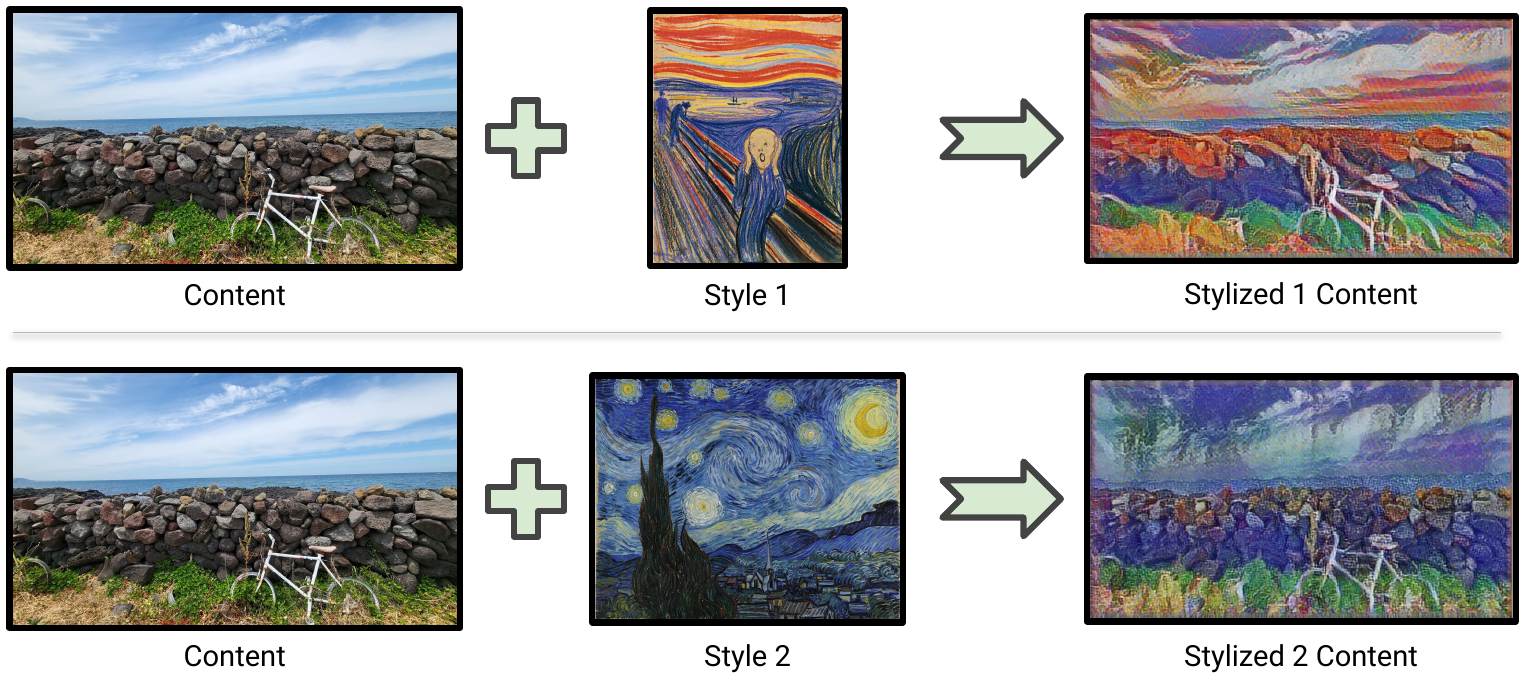}}
	  \newline
    \subfloat[\label{fig:neural_driving_style_transfer}]{	  
        \includegraphics[width=1\columnwidth]{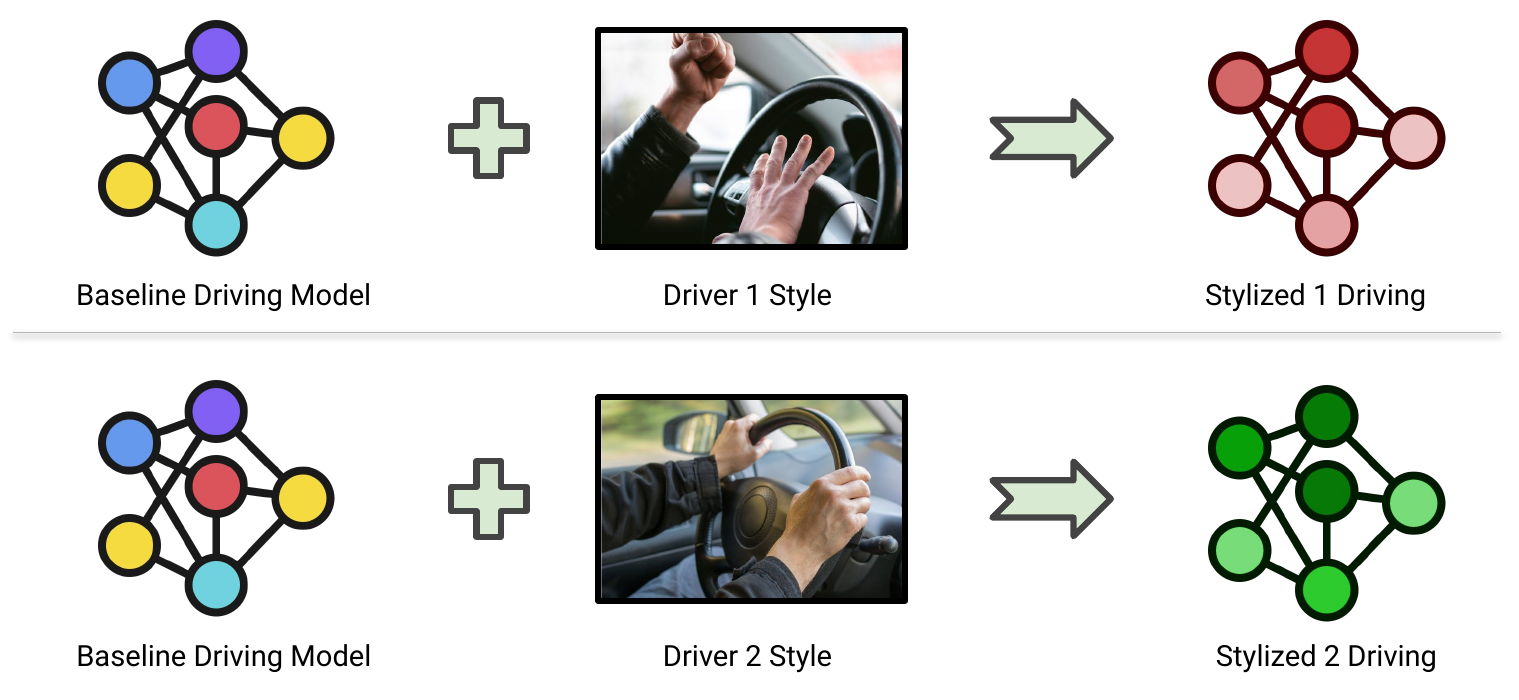}}
    \caption{
    (a) Neural Image Style Transfer (NST) \cite{gatys2015neural}. NST extracts a style from one image and applies the style to the content of another image.
    (b) Neural Driving Style Transfer (NDST). NDST can produce personalized driving results by applying a user's driving style to an existing AD network.
    }
    \label{fig:neural_style_transfer}
\end{figure}

Chandrayee Basu \cite{basu2017you} delves into the investigation of preferred autonomous driving styles, further underscoring this imperative. Basu's research showed that even though the majority of drivers could not accurately identify their own driving style, there was a high correlation between the style they considered their own and the style they preferred, indicating a preference for styles similar to their own. These findings emphasize the need for a personalized human-like AD style.
Providing a personalized human-like AD experience, tailored to match users' unique driving styles while adhering to safety prerequisites, presents a significant opportunity to boost the acceptance of AV \cite{ma2020investigating}\cite{natarajan2022toward}.

\IEEEpubidadjcol 

To achieve this,  we propose a new method specifically designed for vision-based AD. While existing studies \cite{pomerleau_alvinn_1989, net-scale_autonomous_2004, chen_end--end_2017,bojarski2017explaining,wang_end--end_2019-1, wu_end--end_2019, kwon2022incremental, weiss2020deepracing} typically adopt a standardized driving model based on specific applications (e.g., urban driving, racing, etc), our method, \textit{Neural Driving Style Transfer (NDST)}, acknowledges the diversity in drivers' habits influenced by factors like personality and environment \cite{eboli2017drivers}.
Inspired by \textit{Neural Image Style Transfer (NST)} \cite{gatys2015neural} from deep neural networks-based computer vision communities, the proposed method, NDST, leverages the NST concepts to integrate a user's driving style with the vehicle's AD model, creating a personalized driving experience while maintaining standard safety baseline.

The two main components of NDST are the Baseline Driving Model (BDM) and the Personalized Block (PB). Fig.~\ref{fig:neural_style_transfer} illustrates the comparison between NST and NDST. The BDM, serving as the standard AD model, is comparable to the image content in NST, while the PB incorporates the user's driving style. This approach aims to bridge the gap between safety and personalization in AD.

To validate our methodology, we collected data from two drivers with distinct driving styles labeled as; A and B. Style B exhibited faster acceleration and deceleration compared to Style A. The main contributions of this paper are as follows:

\begin{enumerate}
    \item This paper proposes a novel learning method, NDST, designed to provide a human-like personalized AD experience that increases the acceptance of AV and enhances the user experience.
    \item The PB is designed to be a \textit{self-configuring} system, which can be seamlessly integrated with the original AD system (i.e., the BDM) without requiring any modifications. 
    The PB is dependent on the output of the BDM to ensure that any safety-related decisions made by the BDM will be reflected in the final output of the PB. This increases comfort while maintaining the required baseline safety.
    \item An extensive amount of simulations in various test tracks are performed to evaluate the performance of the proposed method. It is observed that the proposed network gives stable autonomous driving on the tracks with the driving style of choice by imitating the driver's acceleration and deceleration patterns.
\end{enumerate}

The remainder of this paper is organized as follows.
Section~\ref{sec:related-work} introduces related research and Section~\ref{sec:method} describes the proposed method.
Section~\ref{sec:results} shows the experimental results in the simulation environment, and analyzes the results.
Section~\ref{sec:conclusion} concludes the thesis and presents directions for future work.

\section{Related Work}
\label{sec:related-work}

Various research efforts have been conducted with the aim to effectively extract and analyze driving styles from drivers' data.

Hironori Hiraishi \cite{hiraishi2021cognitive} used the Neural Image Style Transfer (NST) method to learn the driving style after forming the driving data as an image in RGB format. However, research on how to decode the driving style image to control the vehicle has not been conducted.
Qi et al.\cite{qi2015appropriate} analyzed the driving style based on the change of the temporal distance, which is the distance to the vehicle in front. Based on their analysis, they developed a structure to distinguish different driving styles and to build a connection between style and human driving behavior. However, in their analysis, they do not consider different factors, other than the temporal distance.
Choi et al.\cite{choi2021dsa} proposed a DSA-GAN that recognizes the driver's driving style and predicts the trajectory by reflecting the style. They used Recurrence Plot and CNN network to learn three driving styles, normal, aggressive, and careless, and predicted their trajectories.
In a similar work, Kim et al. \cite{kim2021driving} conducted a work to improve prediction performance and create a personalized trajectory by recognizing and reflecting on various driving styles to predict the trajectory of an ego vehicle. Their proposed system uses a conditional variational autoencoder (CVAE) structure, and studies have been conducted to learn and predict the trajectory of three driving styles: normal, aggressive, and distracted.
However, in these studies, the predicted trajectories were not used to drive the vehicle. This was presented as future work. 
Bingzhao Gao \cite{gao2020personalized} proposed a personalized adaptive cruise control (ACC) system based on model predictive control (MPC). In this work, the driving style was defined as conservative, moderate, and aggressive, and the driving style of the driver was learned using the relative speed and headway of the vehicle in front. 
The work assumed a situation in which a leading vehicle exists, it did not discuss whether the autonomous vehicle continues to adapt its driving style in the absence of a leading vehicle.
Xu et al.\cite{xu2015establishing} proposed a technique for creating driver models that accommodate different driving styles by modeling actual human driving behavior. The authors used old machine learning techniques to examine the driving behaviors of participants who were driving on a closed circuit after collecting their data. The efficiency of the produced driver models was then assessed using a driving simulator. The driving style was represented by the throttle, brake, and vehicle speed data. In another work, Shi et al.\cite{shi2015evaluating} evaluated these driving styles to detect abnormal driving behavior. Similarly, Kuderer et al.\cite{kuderer2015learning} worked on learning the driving style from demonstration, in which they build an algorithm to learn the highway driving style using recorded human driving data. However, these studies were done based on driving data only without visual input as vision-based autonomous driving was not yet popular (imitation learning or behavioral cloning adopting a vision-based end-to-end technique did not acquire popularity until DAVE-2 was developed \cite{chen_end--end_2017}).

\section{Proposed Method}
\label{sec:method}

This section explains how to use the proposed method to transfer the driving style of a driver. The proposed method is based on building a network that learns the driving style of a driver and integrates it into the AV. The driving data used to work the driving style is the acceleration and deceleration data based on road type. However, other driving data can be added as well, such as steering angles, vehicle speed, and time.

The proposed method uses driving data from the user to learn the acceleration/deceleration patterns. To validate our approach, we conducted an experiment using data from two drivers with distinct driving styles, arbitrarily referred to as style A and style B. In practice, the method is designed to accommodate any number of unique driving styles. In our experimental setup, style B represents a driver who accelerates and decelerates more quickly compared to the driver characterized by style A.

\subsection{Neural Driving Style Transfer (NDST) Network Architecture}

\begin{figure*}[!t]
    \centering
    \includegraphics[width=0.99\textwidth]{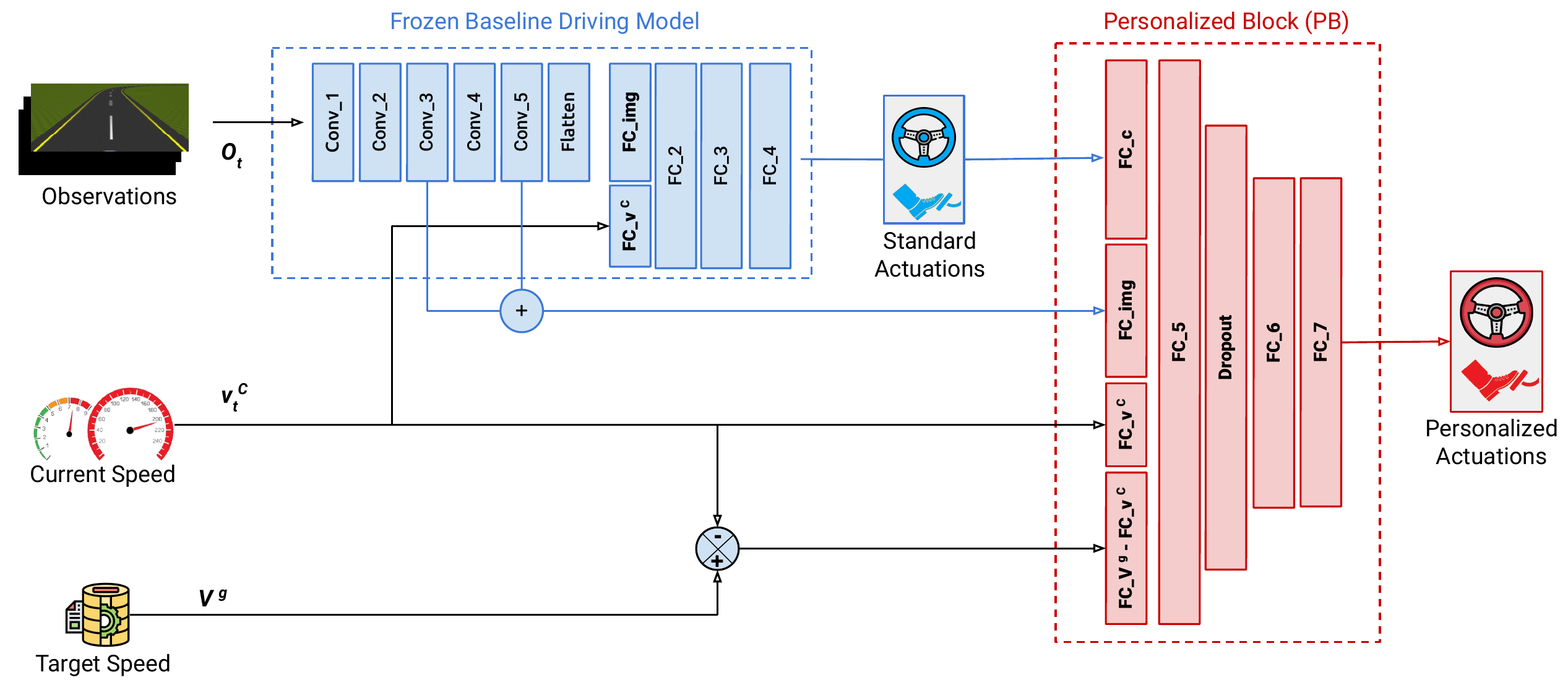}
    \caption{
    Structure of NDST Network.
    The NDST architecture comprises two main components: the Baseline Driving Model (BDM), a standard autonomous vehicle driving model based on NVIDIA's PilotNet, and the Personalized Block (PB), which customizes driving outputs according to individual driver styles. The BDM, depicted in blue, integrates image and speed inputs to generate standard vehicle actuations like steering, throttle, and braking through its five convolutional and three fully-connected layers. The PB, shown in red, inputs the BDM's predicted actions, road features, current speed, and the difference between target and current speeds, outputting personalized steering, throttle, and braking values.
    }
    \label{fig:p-adas_net}
\end{figure*}

The NDST network architecture is depicted in Fig.~\ref{fig:p-adas_net}. The NDST architecture consists of two major blocks: a BDM block and a PB block. The reason behind choosing this design is to make the proposed system a \textit{self-configuring} system, which makes it easy to integrate with the original vehicle system without the need to modify it. 
Therefore, the proposed system, NDST, allows the personalization feature to be easily enabled/disabled based on the user's preference.

\subsubsection{Baseline Driving Model (BDM)}

The baseline driving model is a generic AV driving model that can drive the vehicle.
The BDM network architecture is based on the NVIDIA PilotNet structure \cite{bojarski2016end}, and is shown in blue in Fig.~\ref{fig:p-adas_net}. 
However, the network is modified so that the image and current speed are used as inputs, and the steering angle, throttle, and brake values are defined as outputs.
The network processes the first input (i.e., visual observation) using five convolutional layers, and then the output of the fifth is flattened to form an image feature vector. 
Then, the second input (current speed) is concatenated with the image feature vector.
After that, the concatenated vectors are fed to three fully connected layers (Dense layers). 
The BDM block output is the output of its last fully connected layer and is the prediction of three actuation values (standard actuation in Fig.~\ref{fig:p-adas_net}): Steering, Throttle, and Brake values.

\subsubsection{Personalized Block (PB)}

The PB is a neural network for personalizing the output value of the BDM based on the driving style of a particular driver, which is shown in Fig.~\ref{fig:p-adas_net} in red. It is seen that the PB has four inputs.
The first input is the predicted action values of the BDM. 
The second input is the output of the third and last convolution layer of the BDM after being flattened, which gives an indication of the road conditions.
The third input is the current speed data.
The Fourth input is the difference between the target speed of the vehicle, which is defined by the user and the current speed. 
Each input goes into a fully connected layer to form a vector. 
Then, all four vectors are concatenated and fed to three fully connected layers. 
A dropout layer was needed after the first fully connected layer for generalization purposes. 
The output of the third layer is the final output of the PB which is a personalized predicted actuation based on the style of a particular driver and contains three values: steering angle, throttle, and brake.

\subsection{Training Method for AV Personalization}

\begin{figure}[!t]
    \centering
    \includegraphics[width=1\columnwidth]{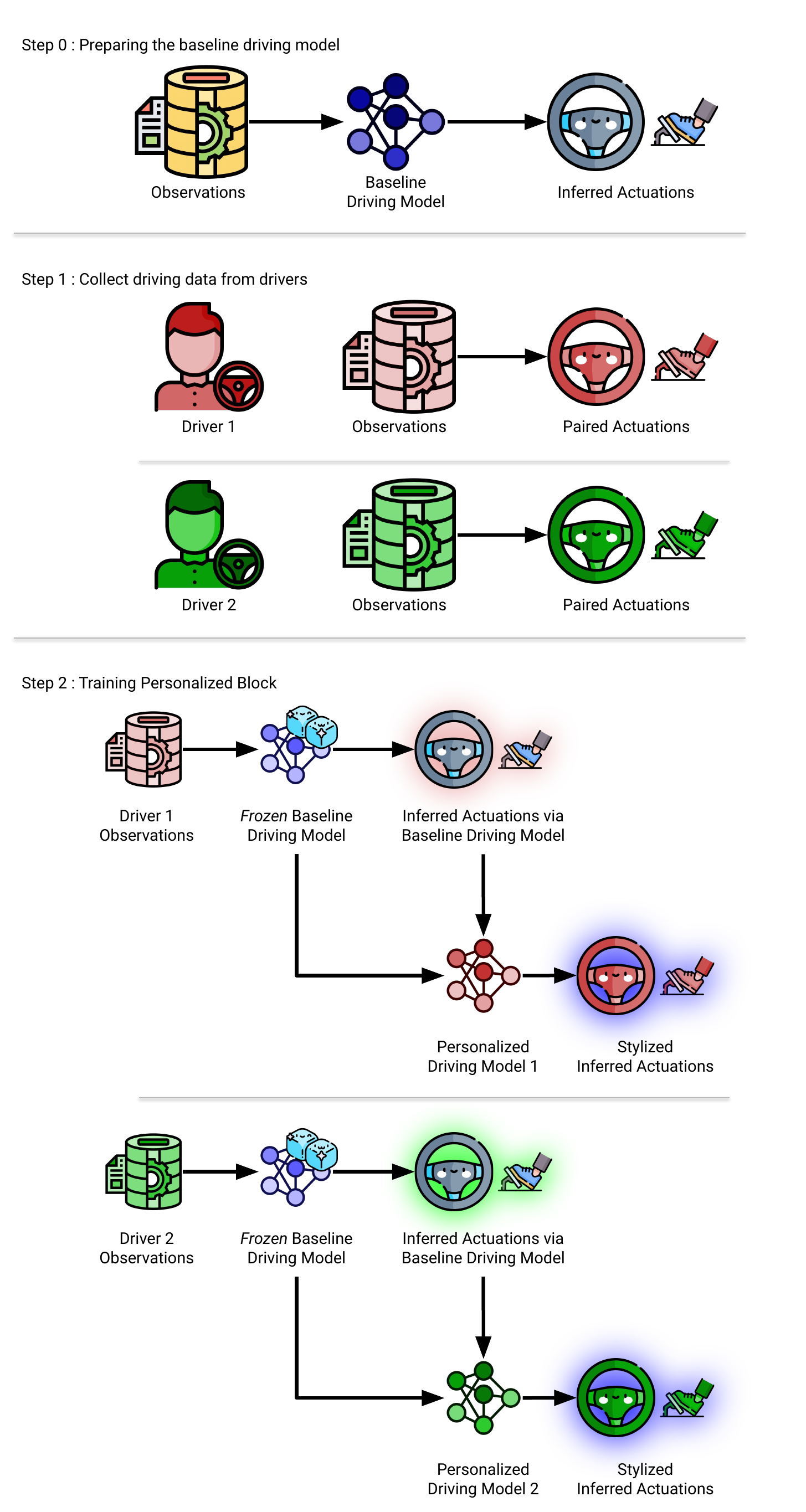}
    \caption{
    Neural driving style transfer (NDST) training process for AVs Personalization.
    The NDST training process begins with establishing a Baseline Driving Model (BDM). Subsequently, the Personalized Block (PB) is trained using data collected for each driver and the BDM, which is set to be non-trainable, ensuring that the PB adjusts and reflects the unique driving style of each driver.
    }
    \label{fig:ndst_system}
\end{figure}

The proposed method for the personalization of AVs is shown in Fig.~\ref{fig:ndst_system}.
The structure of the proposed system consists of the following steps.\\

\noindent\underline{Step 0: Learning the Baseline Driving Model} 

The proposed system essentially works under the presumption that the vehicle is equipped with a baseline autonomy model, called the baseline driving model (BDM), that is capable of driving the vehicle. 
The BDM does not represent the driving preferences of any one driver, thus it is not biased toward the unique driving style of one driver. 

The dataset used to train such a model can be defined as 
\begin{equation}
    D=\{I_t, A_t\}^N_{t=1}, 
    \label{eq:BDM_D}
\end{equation}
where $D$ is dataset, $N$ is the total time steps, $I_t$ is the input data at time $t$ and $A_t$ is the correlative action that controls the vehicle.
The input data $I_t$ can be defined as 
\begin{equation}
    I_{t} = \left\{ O_{t}, v_{t}^{c} \right\},
    \label{eq:BDM_It}
\end{equation}
where $O_{t}$ is visual observation (e.g. images) and $v_{t}^{c}$ is the vehicle's current speed.
The action that controls the vehicle $A_t$ can be defined as
\begin{equation}
    A_{t} = \left\{ S_{t}, T_{t}, B_{t} \right\},
    \label{eq:BDM_At}
\end{equation}
where $S_{t}$, $T_{t}$, and $B_{t}$ are the steering angle, throttle and brake at time $t$, respectively.
Therefore, total input data $I$ and action data $A$ can be defined as
\begin{equation}
    I = \left\{ I_{t}\right\}_{t=1}^{N},\;A = \left\{ A_{t}\right\}_{t=1}^{N}.
    \label{eq:BDM_IA}
\end{equation}

\begin{equation}
    \Tilde{A}=\pi^{BDM} \left ( I \right )
    \label{eq:BDM_prediction}
\end{equation}

In \eqref{eq:BDM_prediction}, $\pi^{BDM}$ represents the BDM policy, which takes $I$ as an input, and provides $\Tilde{A}$ as an output, where $\Tilde{A}$ is the predicted actuation of the BDM.
The BDM is trained to minimize the average prediction error as
\begin{equation}
\omega^* = \underset{\omega}{\arg\min} \sum_{t}^{N} L(\tilde{A}_t, {A}_t),
\label{eq:loss_BDM}
\end{equation}
where $\omega^*$ represents the parameters of the BDM, and L is the loss per sample. 


\noindent\underline{Step 1: Collecting driving data of a driver}

After building the BDM, we start collecting data from a driver whose driving style we want to transfer. This is illustrated in step 1 in Fig.~\ref{fig:ndst_system}, which can be expressed in the following as
\begin{equation}
\begin{split}
    D^n= \{I^{n}, A^{n}\},
\end{split}
\end{equation}
where $n$ represents the driver's number. ${D}^{n}$ is the entire dataset collected by driver $n$. 
${I}^{n}$ and ${A}^{n}$ mean the $I$ and $A$ collected by driver $n$, respectively.


\noindent\underline{Step 2: Training the Personalized Block (PB) for the driver}

Finally, PB is trained using the learned BDM and driving data obtained through Steps 0 and 1.
One PB reflects one driver's style.
In order to train PB only, the baseline driving model is frozen so that it is not trained. 
Input data are then fed to the BDM.
The BDM estimates the 'action' after processing the input data, as shown in \eqref{eq:BDM_prediction}. 
Note that the estimated action is the result of a generic autonomous driving model (i.e., BDM), so non-specific driving style has been applied.

To modify the resulting action into an action that reflects driver $n$'s style, as shown in step 2 in Fig.~\ref{fig:ndst_system}, the PB would get four inputs for training, that are depicted in Fig.~\ref{fig:p-adas_net}.
The first input ($\Tilde{A}^n$) is the predicted action values of the BDM when fed the input data of driver $n$ ($I^n$) is represented by 
\begin{equation}
    \Tilde{A}^{n}=\pi^{BDM} \left ( I^{n}\right ).
    \label{eq:An_Pred_BDM}
\end{equation}
The second input ($\phi(O^{n})$) is obtained by combining the output of two convolution layers when $O^{n}$ is input to BDM, which extracts features from observation data ($O^n$). This data is used to give the PB spatial information about the road.
The third input is the current speed data collected by driver $n$ in step 1 $\left ( v^{c} \right )^{n}$.
The last input is the difference between the target speed of the vehicle ($V^g$) which is defined by the user and the current speed $v^{c}$. 
The target speed is used by PB to determine acceleration and deceleration.
The action obtained through driver $n$ ($A^n$) is used as the ground truth for learning.
The final output of the trained PB is a personalized predicted actuation of driver $n$ ($\Tilde{\Tilde{{A}}}^{n}$) and can be defined as
\begin{equation}
    \Tilde{\Tilde{A}}^{n}=\pi^{PB}\left ( \Tilde{A}^{n}, \phi \left ( O^{n} \right ) , \left ( v^{c} \right )^{n}, V^{g} \right ).
    \label{eq:An_Pred_PB}
\end{equation}
The loss function used to learn PB can be defined as
\begin{equation}
\chi^* = \underset{\chi}{\arg\min} \sum_{t}^{N} L(\Tilde{\Tilde{A}}_{t}^{n}, {A}_{t}^{n}),
\label{eq:loss_PB}
\end{equation}
where $\chi$ represents parameters of the PB, and $L$ is a loss per sample.

The proposed method for personalization is designed not to modify the original system but to add to it, by plugging the PB block into the vehicle system. 

\subsection{Simulation Environment}

To implement the NDST, we used the OSCAR simulator \cite{kwon_oscar_2021} which was introduced in \cite{kwon2022incremental}, for its simplicity as it was built using ROS (Robotic Operating System) \cite{quigley2009ros} and Gazebo simulator \cite{koenig2004designGazebo} and because it was designed for vision-based autonomous driving.

\subsubsection{Track}

\begin{figure}[!t]
    \centering
    \includegraphics[width=1\columnwidth]{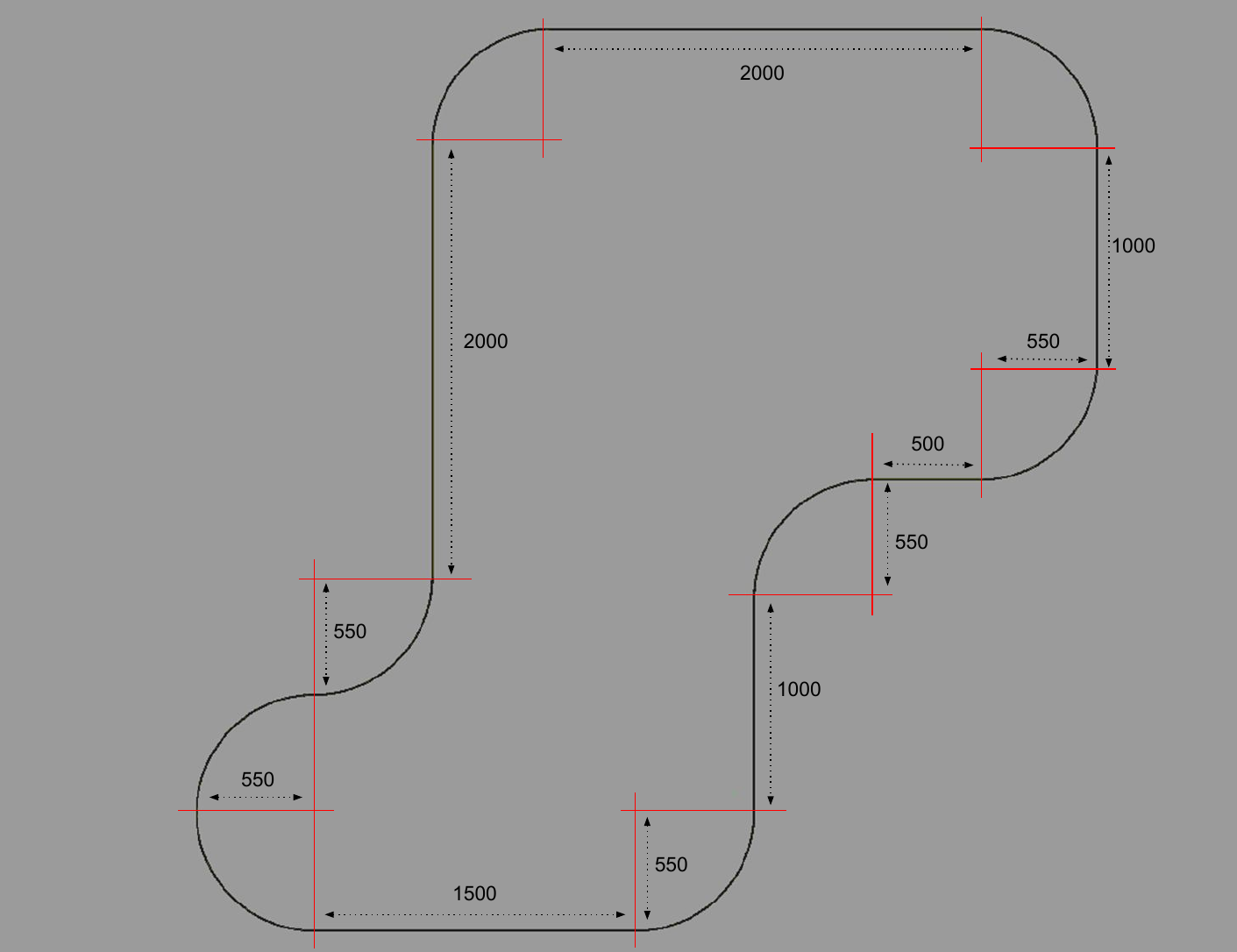}
    \caption{
    The training track used to collect training data from two different drivers. 
    }
    \label{fig:track_train}
\end{figure}
\begin{figure}[!t]
    \centering
    \includegraphics[width=1\columnwidth]{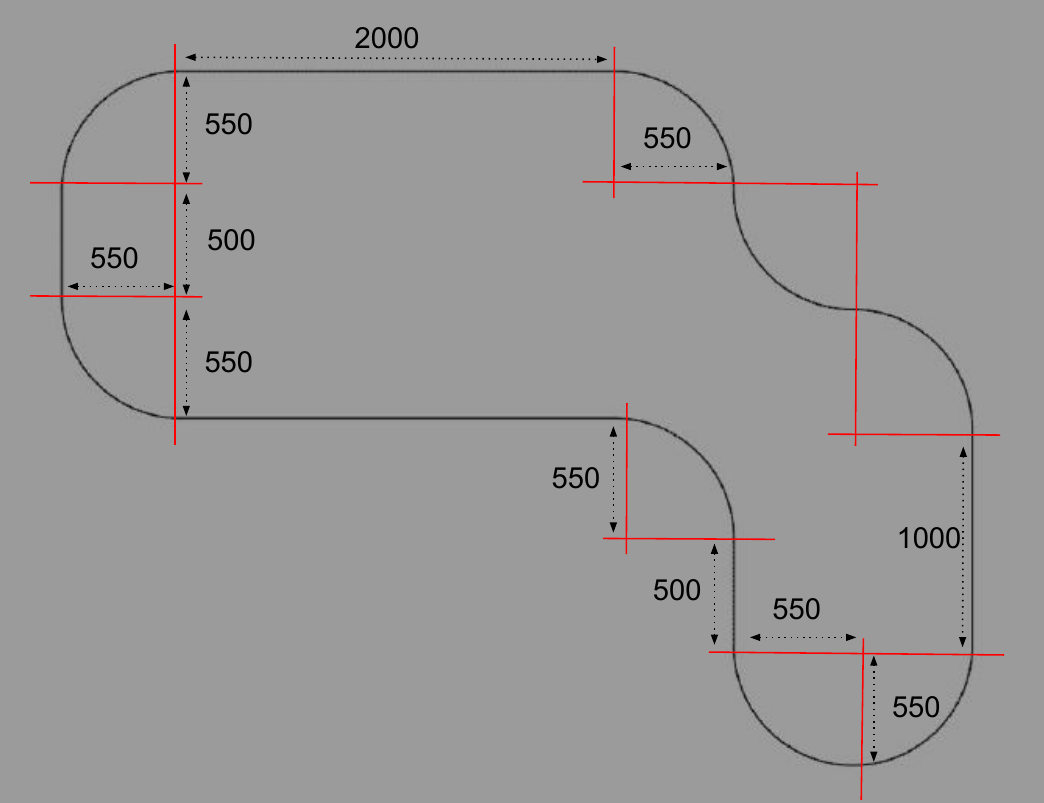}
    \caption{
    The test track used to test Neural Driving Style Transfer (NDST) system.
    }
    \label{fig:track_test}
\end{figure}

Since the driving style we are trying to transfer is represented by the steering angle, throttle (acceleration), and brake (deceleration), the training track, shown in Fig.~\ref{fig:track_train}, was designed to cover three cases that reflect the driving style: acceleration/deceleration to reach the target speed, deceleration before entering the curved section, and acceleration after exiting the curved section. To test the NDST model after training, a test track was constructed to check the acceleration and deceleration of straight and curved sections, and the corresponding track is shown in Fig.~\ref{fig:track_test}.

\begin{table}[!t]
\caption{Total number of data collected for training.}
\label{tbl:dataset}
\centering
\begin{tabular}{|l|l|l|}
\hline
\textbf{Train model}            & \textbf{Driving style} & \textbf{Number of Data} \\ \hline 
Baseline Driving Model & Standard      & \multicolumn{1}{c|}{85910}          \\ \hline
Personalized Block 1   & Style A  & \multicolumn{1}{c|}{41321}          \\ \hline
Personalized Block 2   & Style B    & \multicolumn{1}{c|}{37673}          \\ \hline
\end{tabular}
\end{table}
\begin{figure}[!t]
    \centering
    
    \subfloat[Style A Straight\label{fig:con_straight}]{%
	    \includegraphics[width=0.33\columnwidth]{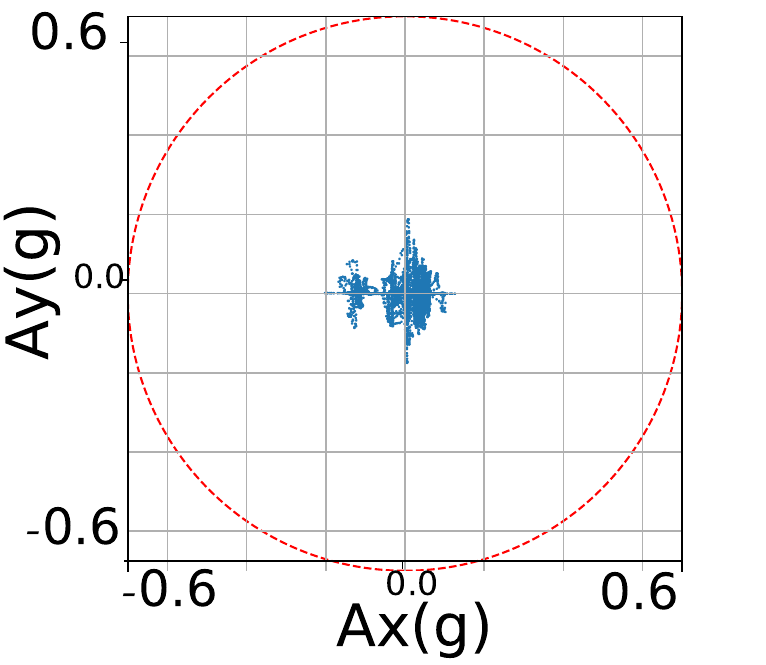}}
        \hfill
    \subfloat[Style A Left\label{fig:con_left}]{%
	    \includegraphics[width=0.33\columnwidth]{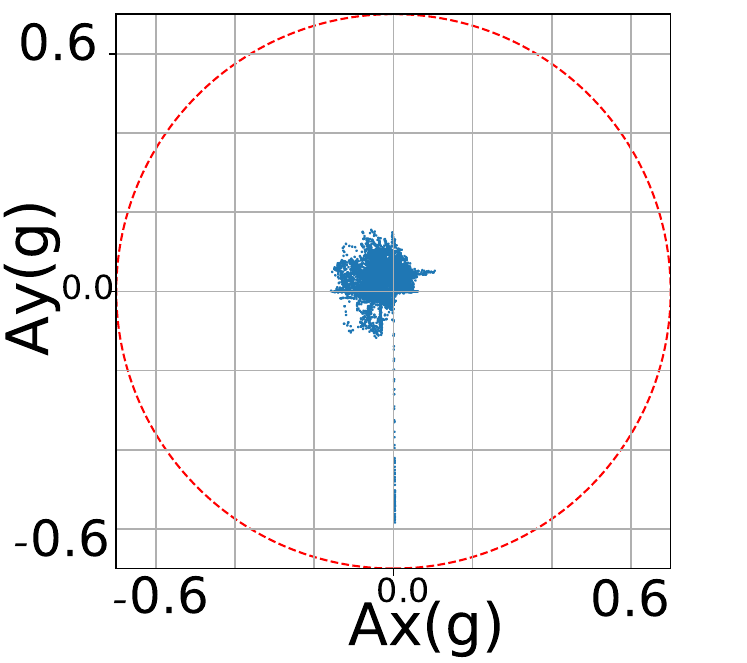}}
	    \hfill
    \subfloat[Style A Right\label{fig:con_right}]{%
        \includegraphics[width=0.33\columnwidth]{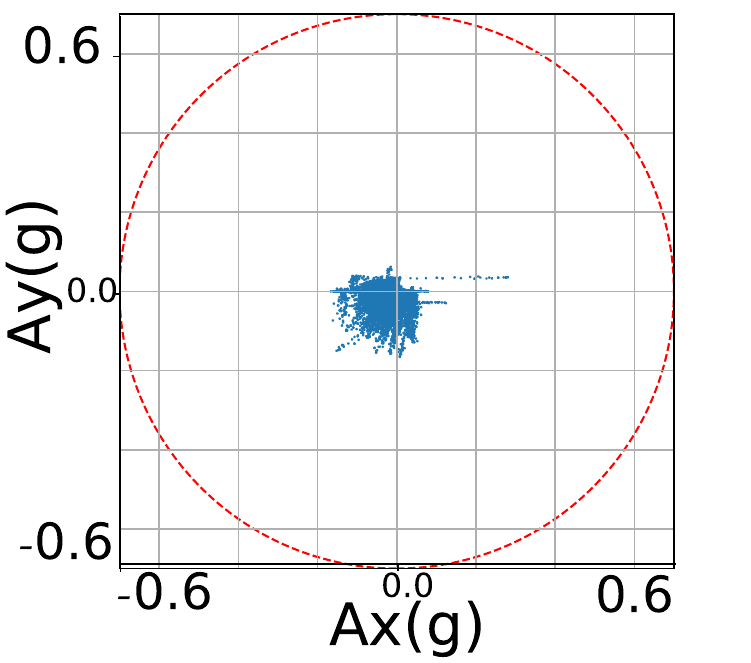}}
	    \hfill
	  \newline
    \subfloat[Style B Straight\label{fig:agg_straight}]{%
	    \includegraphics[width=0.33\columnwidth]{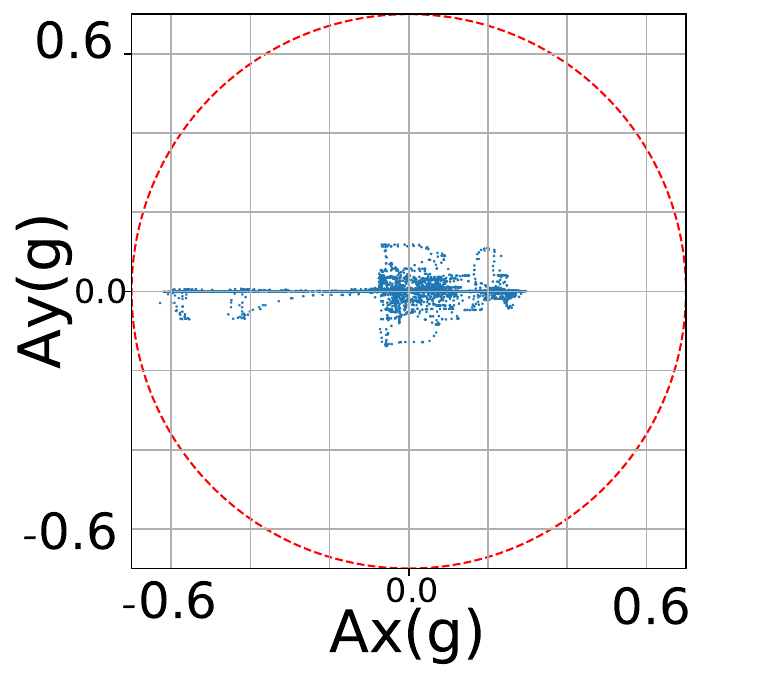}}
        \hfill
    \subfloat[Style B Left\label{fig:agg_left}]{%
	    \includegraphics[width=0.33\columnwidth]{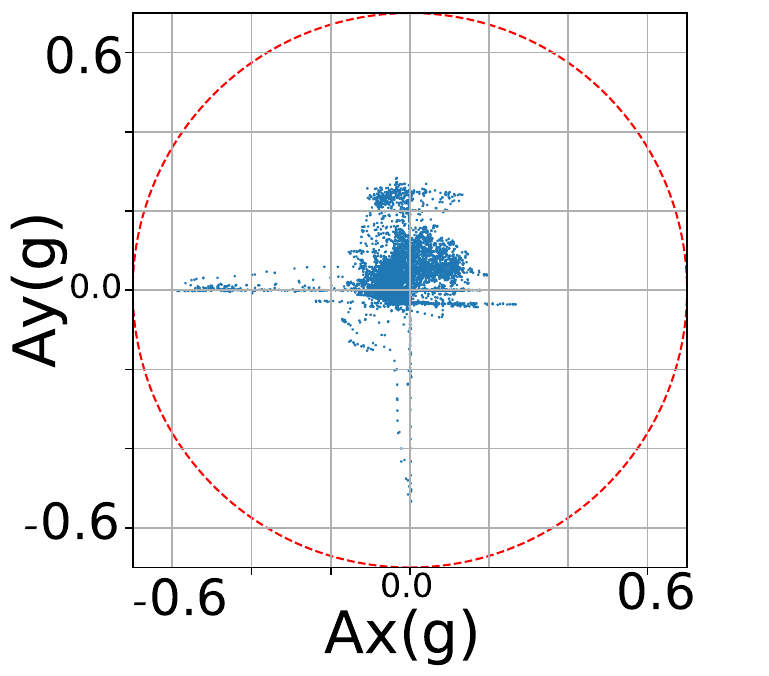}}
	    \hfill
     \subfloat[Style B Right\label{fig:agg_right}]{%
        \includegraphics[width=0.33\columnwidth]{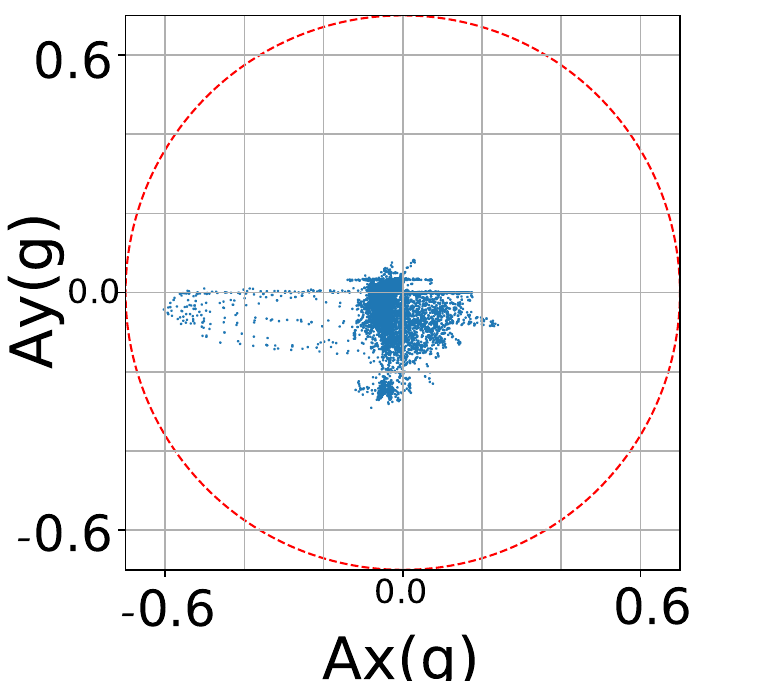}}
    \caption{
    (a), (b), and (c) each represent the G-G diagram of the straight, left-turn, and right-turn roads in Style A, respectively. Similarly, (d), (e), and (f) each correspond to the G-G diagram of the straight, left-turn, and right-turn roads in Style B, respectively.
    }
    \label{fig:dataset_gg}
\end{figure}

\begin{figure}[!thb]
    \centering
    \subfloat[\label{fig:plot_styleA_training}]{
	    \includegraphics[width=1\columnwidth]{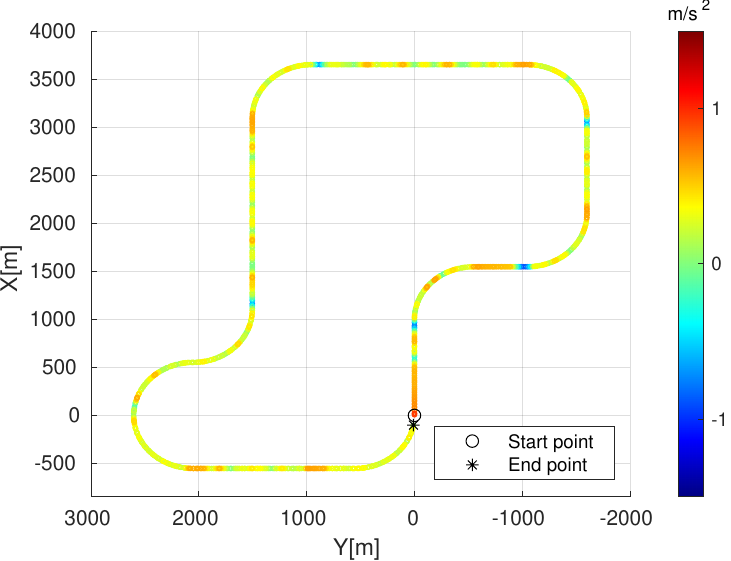}}   
	  \hfill
    \subfloat[\label{fig:plot_styleB_training}]{
	    \includegraphics[width=1\columnwidth]{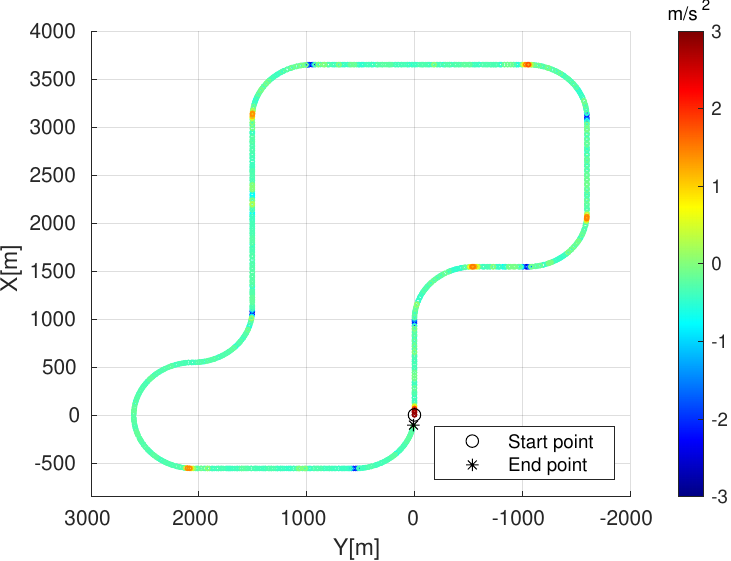}} 
    \caption{
    Training dataset visualization on the training track. (a) The dataset was collected by driver 1 (driving in style A) on the training track.
    (b) The dataset was collected by driver 2 (driving in style B) on the training track.
    }
    \label{fig:plot_style_training}
\end{figure}

\subsubsection{Dataset}
Before training the NDST Network, the BDM should be trained to enable normal driving without reflecting a specific driving style. Therefore, in order to collect a dataset suitable for this purpose and eliminate any bias in the data, the BDM was trained using multiple datasets with various acceleration, deceleration, and speed data on the track.
Afterward, training data was collected to train two personalized blocks (PB1 and PB2), representing driving data in two different styles; style A and style B. 
Table \ref{tbl:dataset} shows the number of data collected for training the BDM and the two personalized blocks. The total number of training data was $85.9$k, $41.3$k, and $37.6$k, for BDM, PB1, and PB2, respectively.

In style A, the rate of change in acceleration and deceleration was low compared to style B.
The G-G diagram visualizing the acceleration of style A and style B data is shown in Fig.~\ref{fig:dataset_gg}. 
On the straight road, the style A G-G diagram can confirm that the range of x-axis acceleration is relatively smaller than the style B G-G diagram, and the area of the G-G diagram measured for the left and right turn roads is also smaller than style B.
This shows that style B exhibits more rapid acceleration and deceleration than style A in terms of vehicle control.
As shown in Fig.~\ref{fig:plot_style_training}, both styles decelerate upon entering the curve and accelerate upon exiting the curve, but the values of the acceleration and deceleration are different for each style. 
Style B (Fig.~\ref{fig:plot_styleB_training}) had higher values of acceleration and deceleration compared to style A (Fig.~\ref{fig:plot_styleA_training}), and acceleration and deceleration were controlled quickly and over a shorter distance.

    \section{Experimental Results}
\label{sec:results}

In our experiments, we used a variety of plots to validate the performance of the NDST method.
These evaluations confirmed that \textit{self-configuring} PB can effectively transfer a driver's driving style into a generic AV model for personalization.

In the training phase, the NDST method was used to learn driving styles, style A and style B. During this process, the PBs learned the distribution of acceleration and deceleration control values associated with each driving style.
\begin{figure}[!tb]
    \centering
    \subfloat[PB1 (transfer to Style A)\label{fig:plot_style_gg_con}]{
	    \includegraphics[width=0.45\columnwidth]{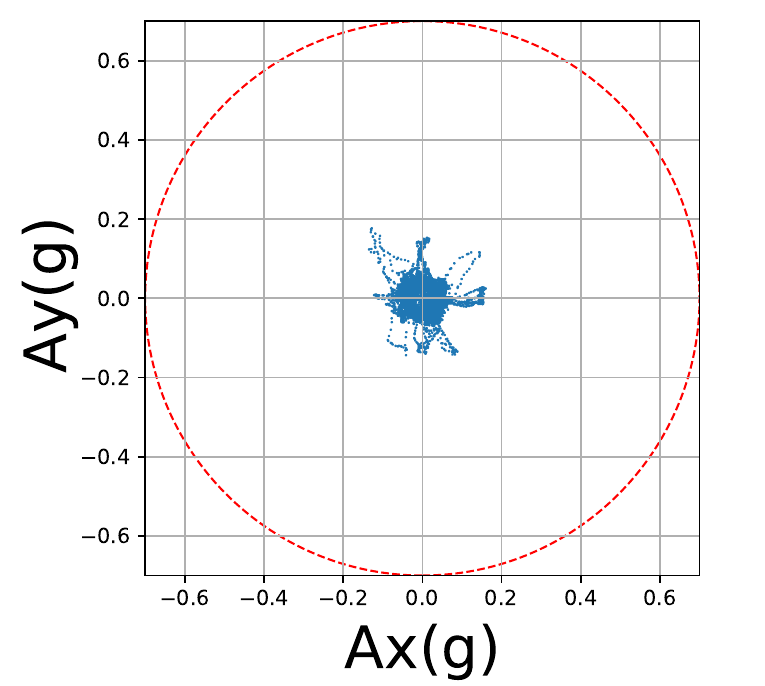}}
     \hfill
    \subfloat[PB2 (transfer to Style B)\label{fig:plot_style_gg_agg}]{
	    \includegraphics[width=0.45\columnwidth]{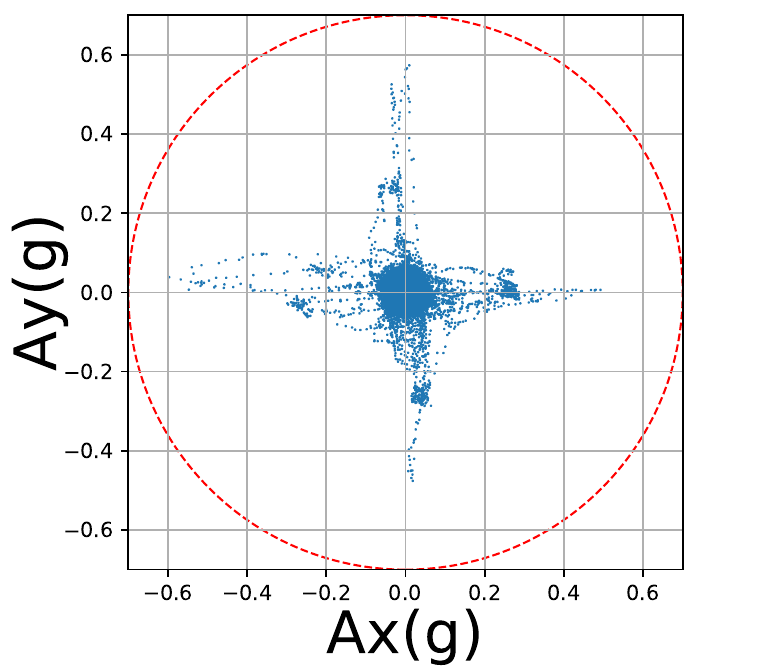}}
    \caption{
    Results from driving on the test track with trained PBs (a) A G-G diagram of the acceleration/deceleration results from PB1 which transferred Style A.
    (b) a G-G diagram of the acceleration/deceleration results from PB2 which transferred Style B.
    }
    \label{fig:plot_style_gg_model}
\end{figure}

The training results were evaluated by analyzing acceleration and deceleration outputs obtained from test track drives using PB1 and PB2, respectively. Subsequent verification confirmed the successful execution of the style transfer.
The result is seen in Fig.~\ref{fig:plot_style_gg_model}, where Fig.~\ref{fig:plot_style_gg_con} is a G-G diagram of the acceleration/deceleration results from PB1 which transferred style A, and Fig.~\ref{fig:plot_style_gg_agg} is a G-G diagram of the acceleration/deceleration results from PB2 which transferred style B.
The difference between the two G-G diagrams clearly shows that the styles have been transferred successfully, as each PB drove in a different style.
PB1 does not have a large change in acceleration, so the area displayed on the G-G diagram is small. 
In contrast, PB2 (which transfers style B) 
has a large change in acceleration, almost three times the acceleration value compared to style A transferred by PB1.

\begin{figure}[!t]
    \centering
    \subfloat[\label{fig:plot_styleA_testing}]{
	    \includegraphics[width=1\columnwidth]{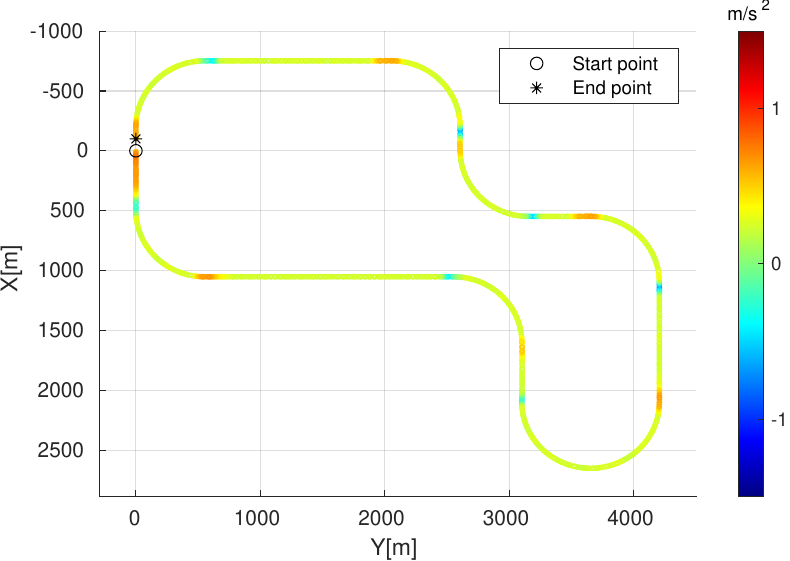}} 
	  \hfill
    \subfloat[\label{fig:plot_styleB_testing}]{
	    \includegraphics[width=1\columnwidth]{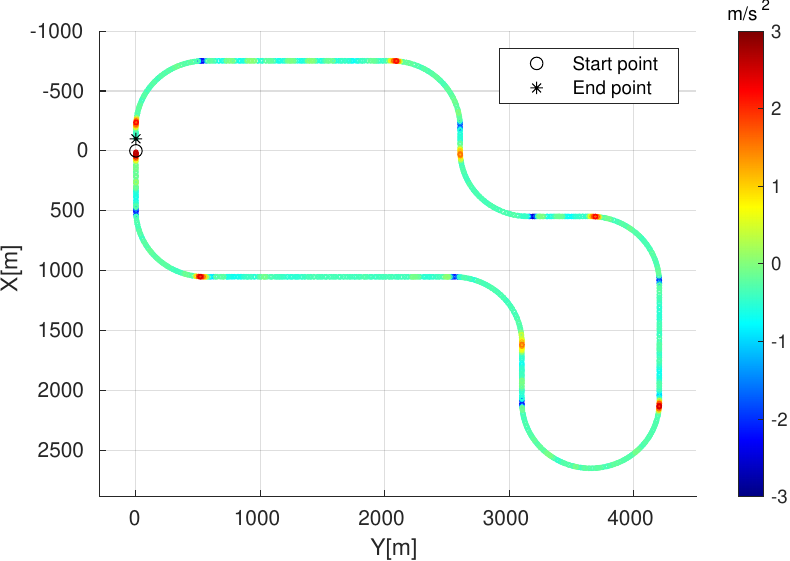}}  
    \caption{
    Testing driving data.
    (a) Style A driving data. Trained NDST model with PB1 driving on the test track.
    (b) Style B driving data. Trained NDST model with PB2 driving on the test track.
    }
    \label{fig:plot_style_testing}
\end{figure}


\begin{figure}
  \centering
  \usetikzlibrary{spy,backgrounds}
  \begin{tikzpicture}[spy using outlines]  
    \node (x)  {\includegraphics[width=0.47641\textwidth]{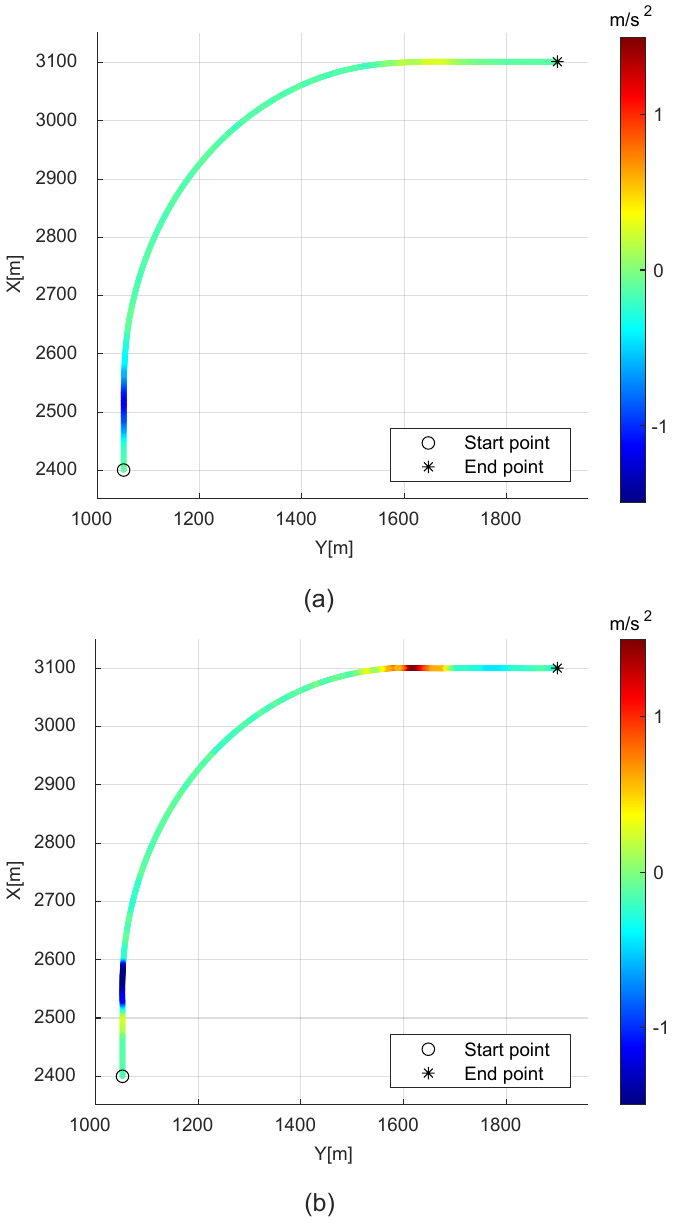}};
  \begin{pgfonlayer}{background}
    
   \spy[rectangle, black, magnification=2, size=2.6cm] on (1, 6.8) in node at (1, 4.1);
  
  \spy[rectangle, black, magnification=2, size=2.6cm] on (0.8, -0.8) in node at (1, -3.5);
  \end{pgfonlayer}
  \draw[black] (0.9,6.15)--(1,5.4);
  \draw[black] (0.8,-1.45)--(1,-2.2);
\end{tikzpicture}

\caption{Acceleration/deceleration in curved sections
(a) Results from PB1 sending Style A.
(b) Results from PB2 sending Style B.}
  \label{fig:short_track}
\end{figure}
\begin{figure*}[!t]
    \centering
    \subfloat[\label{fig:distance_result_straight_A}]{
        \includegraphics[width=1.934\columnwidth]{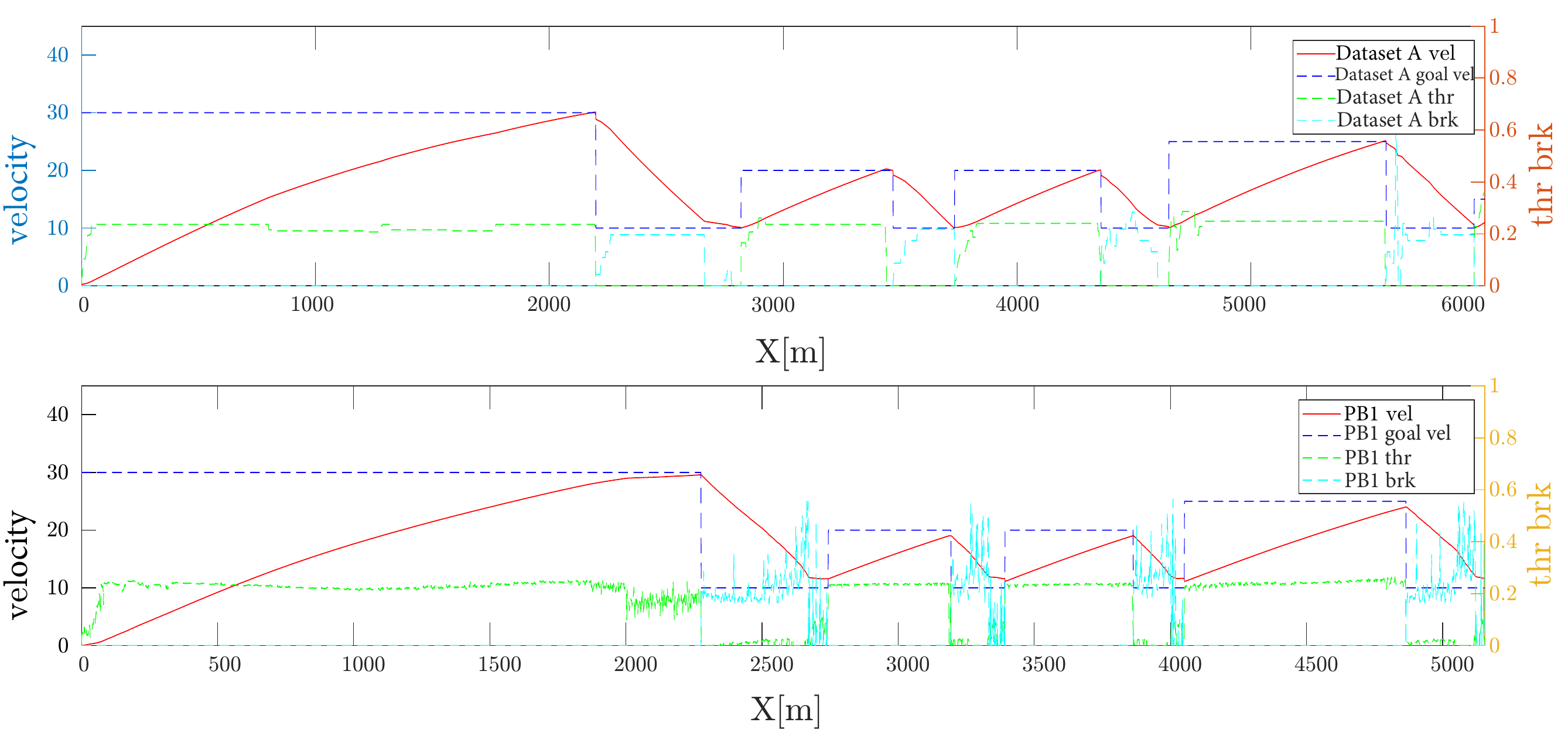}}
    \hfill
    \subfloat[\label{fig:distance_result_straight_B}]{
    \includegraphics[width=1.934\columnwidth]{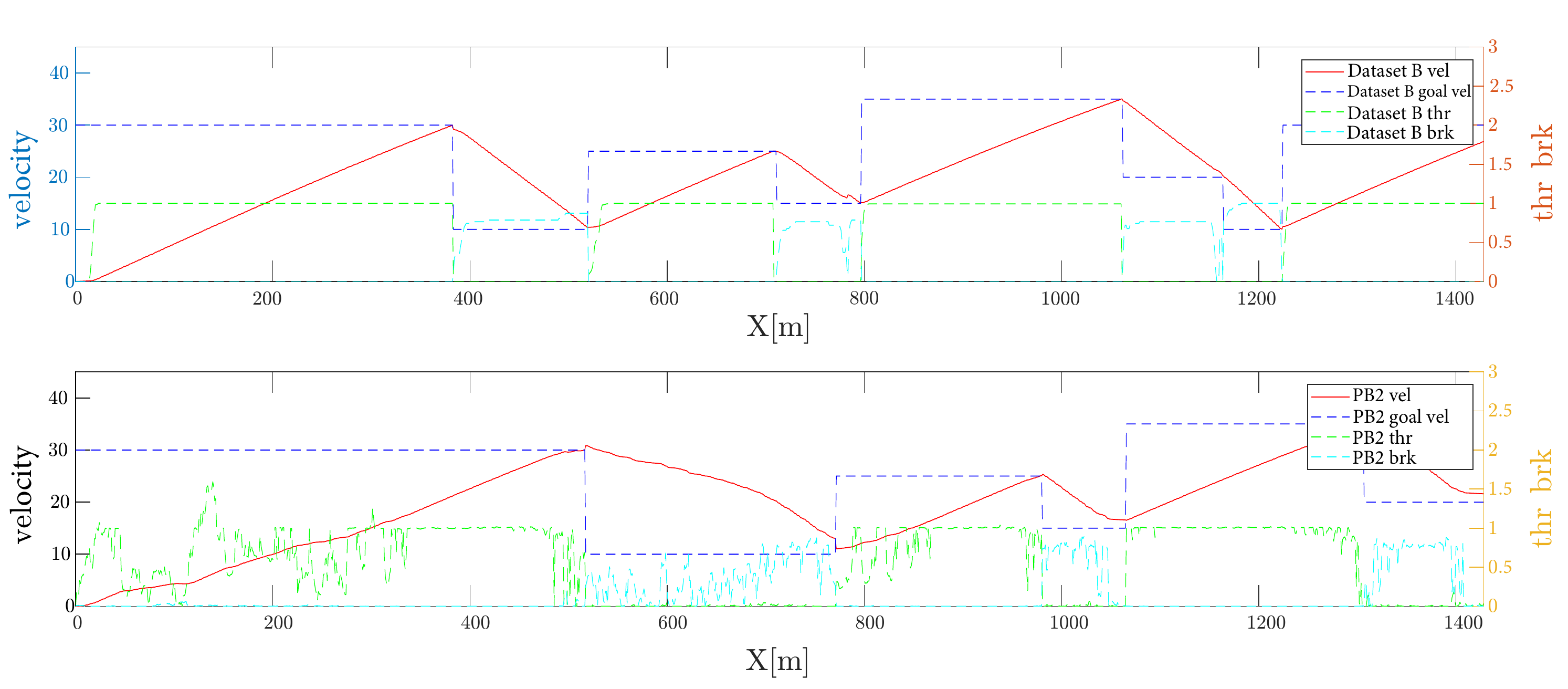}}
    \caption{
    The distance needed for each model to reach a target speed of $30 m/s$ in a straight road section.
    (a) The driving data of the original dataset was collected by driver 1 (top) and the PB1 (style A) driving data (bottom).
    (b) The driving data of the original dataset was collected by driver 2 (top) and the PB2 (style B) driving data (bottom). 
    The PB1 (style A) took $2250 m$ from the starting point to reach $30 m/s$, but the PB2 (style B) reached $30 m/s$ in $500 m$, showing rapid acceleration and deceleration.
    }
    \label{fig:distance_result_straight}
\end{figure*}

Based on the results of driving on the test track, using a 3D plot, the acceleration values for each road section were visualized. 
Fig.~\ref{fig:plot_style_testing} is the result of visualizing these acceleration values.
Fig.~\ref{fig:plot_styleA_testing} shows the result of driving on the test track in style A using the trained PB1, and Fig.~\ref{fig:plot_styleB_testing} shows the result of driving on the same test track in style B using the trained PB2.
Since acceleration/deceleration values are higher in style B, we can see that the acceleration/deceleration periods of time are shorter in comparison with style A. This is obvious at the points of entering or exiting a curved road section, as depicted in Fig.~\ref{fig:short_track}.

The target speed is set to $20$m/s and shows the vehicle accelerates from the starting point to reach the target speed. 
It then shows the car slowing down as it enters a curve and accelerating as it exits.
While on the curve itself, both PB1 and PB2 slowed down from $20$m/s to $15$m/s before entering the curve, driving through the curve and maintaining $15$m/s. While reaching the curve exit, we compared the acceleration and distance required by the two models to reach the target speed of $20$m/s after escaping from the curve.
Although both PB1 and PB2 accelerate and decelerate to reach the target speed, the acceleration, and deceleration applied to reach that target speed are different.
PB2 (style B) reached the target speed of $80$m, while PB1 traveled $150$m to reach the target speed. 
Fig.~\ref{fig:distance_result_straight} shows the velocity and control values of each model when the target speed is set on the straight road section.
Fig.~\ref{fig:distance_result_straight_A} shows the driving data of the original dataset collected by driver 1 (top) and the PB1 (style A) driving data (bottom), and Fig.~\ref{fig:distance_result_straight_B} shows the driving data of the original dataset collected by driver 2 (top) and the PB2 (style B) driving data (bottom). 
The PB1 (style A) took $2250$m from the starting point to reach $30$m/s, but the PB2 (style B) reached $30$m/s in $500$m, showing rapid acceleration and deceleration.


    \section{Conclusion}
\label{sec:conclusion}
{
This paper introduces a novel approach called Neural Driving Style Transfer (NDST) that aims to personalize the driving experience of vision-based AV. 
NDST can transfer a driver's unique driving style to a vision-based AV, allowing the vehicle to drive in a way that is more similar to the driver's preferences by integrating a standardized driving model with a Personalized Block (PB) capable of acquiring causal associations between acceleration and deceleration from each driver's unique data. 
The proposed system, NDST, was successfully validated through simulated driving results, underscoring its ability to adapt an AV system to individual driving styles.
}

    \section*{Acknowledgment}

    \bibliographystyle{IEEEtran}
\bibliography{IEEEabrv, references}

    \begin{IEEEbiography}
    [{\includegraphics[width=1in,height=1.25in,clip,keepaspectratio]{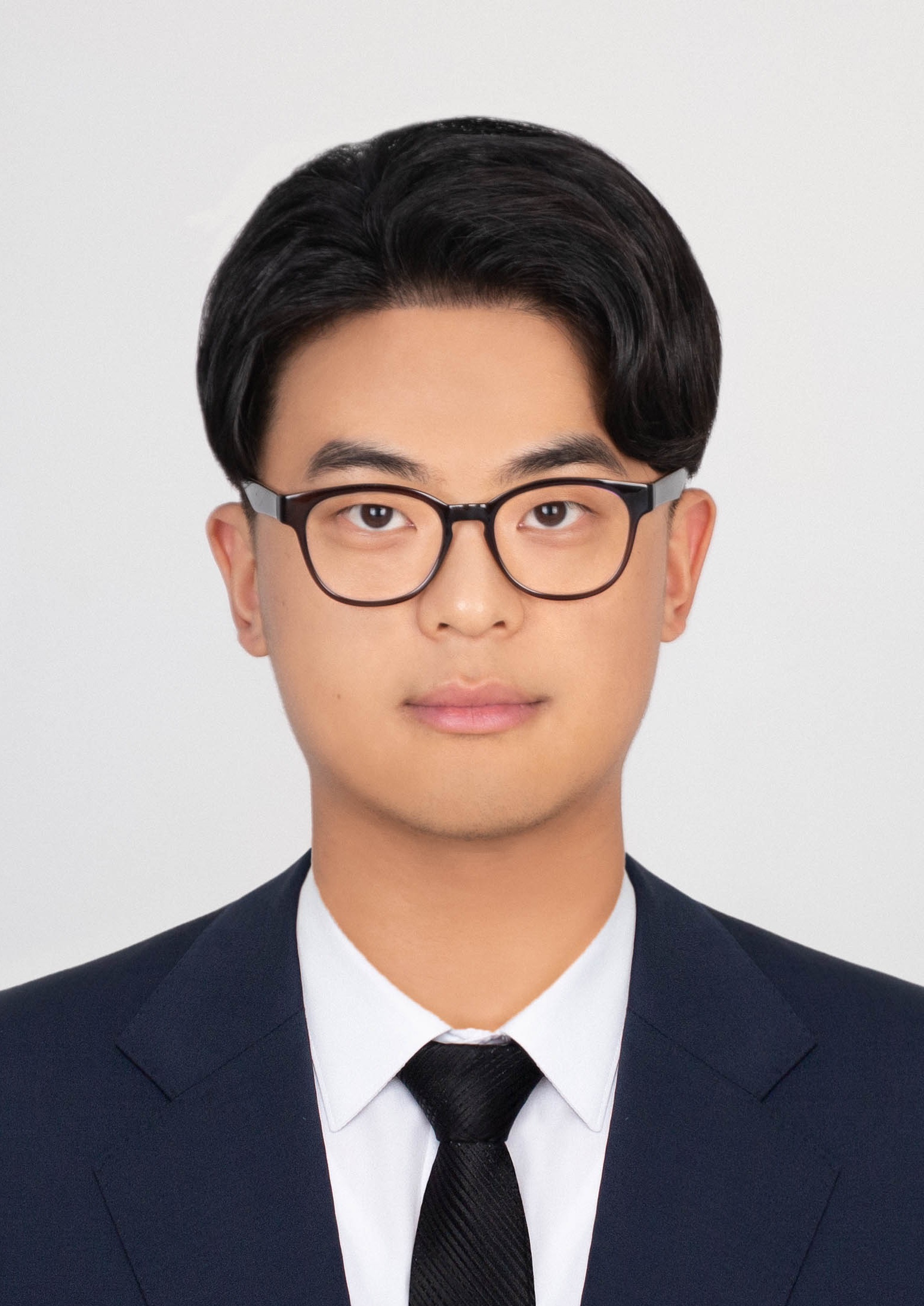}}]
	{Donghyun Kim} received the B.S. degree in electronics and communication engineering from Hanyang University, Ansan, South Korea, in 2018, where he is currently pursuing the Ph.D. degree. His research interests include autonomous vehicles, personalized autonomous driving, and artificial intelligence. He received numerous scholarships, such as the Korean Government Scholarship for his M.S. and Ph.D. degrees.
\end{IEEEbiography}

\begin{IEEEbiography}
    [{\includegraphics[width=1in,height=1.25in,clip,keepaspectratio]{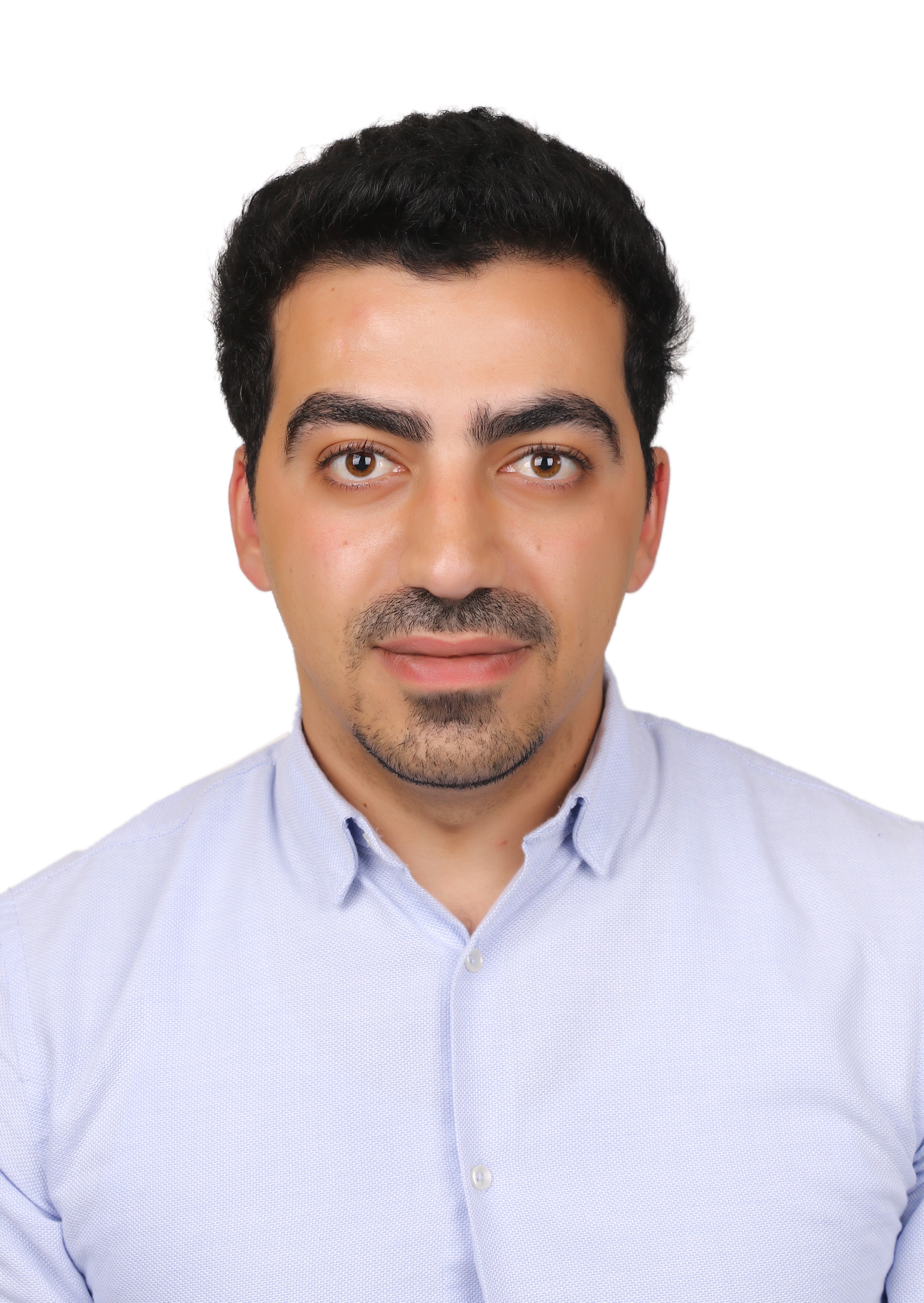}}]
    {Aws Khalil} earned a B.S. in Mechatronics Engineering from the University of Jordan, Amman, Jordan in 2018. He received his M.S. degree from Budapest University of Technology and Economics (BME) in Autonomous Vehicle Control Engineering. He is currently a Ph.D. student in Electrical, Electronics, and Computer Engineering at the University of Michigan - Dearborn. Mr. Khalil is interested in autonomous vehicles, and machine learning. During his master's studies, he worked as a research assistant at BME Automated Drive Lab, and now he works at the University of Michigan - Dearborn's Bio-inspired Machine Intelligence (BIMI) Laboratory.
\end{IEEEbiography}

\begin{IEEEbiography}
    [{\includegraphics[width=1in,height=1.25in,clip,keepaspectratio]{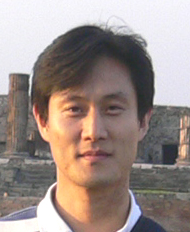}}]
	{Haewoon Nam} (S’99–M’07–SM’10) received the B.S. degree from Hanyang University, Seoul, South Korea, the M.S. degree from Seoul National University, and the Ph.D. degree in electrical and computer engineering from The University of Texas at Austin, Austin, TX, USA. From 1999 to 2002, he was with Samsung Electronics, Suwon, South Korea, where he was engaged in the design and development of code division multiple access and global system for mobile communications (GSM)/general packet radio service baseband modem processors. In the summer of 2003, he was with the IBM Thomas J. Watson Research Center, Yorktown Heights, NY, USA, where he performed extensive radio channel measurements and analysis at 60 GHz. In the fall of 2005, he was with the Wireless Mobile System Group, Freescale Semiconductor, Austin, where he was engaged in the design and test of the worldwide interoperability for microwave access (WiMAX) medium access control layer. His industry experience also includes work with the Samsung Advanced Institute of Technology, Kiheung, South Korea, where he participated in the simulation of multi-input–multi-output systems for the Third-Generation Partnership Project (3GPP) Long-Term Evolution (LTE) standard. In October 2006, he joined the Mobile Devices Technology Office, Motorola, Inc., Austin, where he was involved in algorithm design and development for the 3GPP LTE mobile systems, including modeling of 3GPP LTE modem processor. Later in 2010, he was with Apple Inc., Cupertino, CA, USA, where he worked on research and development of next-generation smart mobile systems. Since March 2011, he has been with the Division of Electrical Engineering, Hanyang University, Ansan, South Korea, where he is currently a Professor. He received the Korean Government Overseas Scholarship for his doctoral studies in the field of electrical engineering.
\end{IEEEbiography}

\begin{IEEEbiography}
    [{\includegraphics[width=1in,height=1.25in,clip,keepaspectratio]{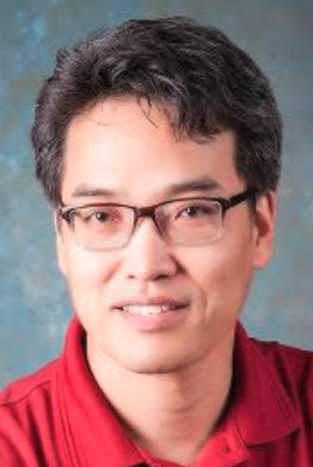}}]
    {Jaerock Kwon} (M'06--SM'20) received the B.S. and M.S. degrees from Hanyang University, Seoul, South Korea, in 1992 and 1994, respectively, and the Ph.D. degree in computer engineering from Texas A\&M University, College Station, USA, in 2009. Between 1994 and 2004, he worked at LG Electronics, SK Teletech, and Qualcomm Internet Service. From 2009 to 2010, he was a Professor with the Department of Electrical and Computer Engineering, Kettering University, Flint, MI, USA. Since 2010, he has been a Professor with the Department of Electrical and Computer Engineering, University of Michigan-Dearborn, MI, USA. His research interests include mobile robotics, autonomous vehicles, and artificial intelligence. Dr. Kwon’s awards and honors include the Outstanding Researcher Award, Faculty Research Fellowship (Kettering University), and SK Excellent Employee (SK Teletech). He serves as the President of the Korean Computer Scientists and Engineers Association in America (KOCSEA), in 2020 and 2021.
\end{IEEEbiography}

\end{document}